\documentstyle{article}

\title{Empirical Discovery in Linguistics}

\author{Vladimir Pericliev\\
Department of Mathematical Linguistics\\
Institute of Mathematics with Computing Centre, bl.8\\
Bulgarian Academy of Sciences, 1113 Sofia, Bulgaria\\
e-mail: peri@bgearn.acad.bg}

\begin{document}

\maketitle

\begin{abstract}
A discovery system for detecting correspondences in data is described,
based on the familiar induction methods of J.\ S.\ Mill. Given a set of
observations, the system induces the ``causally'' related facts in these
observations.  Its application to empirical linguistic discovery is
described.  The paper is organized as follows.  I begin the discussion by
revealing two developments, the transformationalists' critique of ``discovery
procedures'' and naive inductivism, which have led to the neglect of discovery
issues, arguing that more attention needs to be paid to discovery in
linguistics.  Then, Mill's methods are introduced, and the system,
incorporating them, is described, using as one illustration the discovery of (a
part of) the famous Germanic Consonant Shift, known as Grimm's Law.
\end{abstract}

\section{Introduction}

Scientific discovery was one of the favourite topics of Renaissance
scholars (F.  Bacon, Descartes, Leibnitz).  These early efforts suffered a
long period of oblivion (basically, due to the critiques of Whewell and
Hume), but this century has witnessed a steady, though perhaps not a
completely uninterrupted, \footnote{I have in mind esp.  the banning of
discovery from the realm of a science like philosophy until the late fifties
(cf.  Hanson 1958), triggered by Popper's denunciation of a logic of
discovery in his classic {\em The Logic of Scientific Discovery} (German
original {\em Logik der Forschung} from 1935); for a discussion, cf.\ e.g.
Simon 1973, Nickles 1980a. } process of revival of interest.  Very significant
contributions to a general understanding of the discovery process have come
from various scientific disciplines, incl.\ mathematics (Poincar\'e, Hadamard,
Polya), psychology (Wertheimer, Duncker), philosophy (Nickles 1980b, Nickles
1980c), and AI (Newell and Simon 1972, Langley, Simon, Bradshaw and \.Zytkow
1987); cf.  also the collection of more recent advances (Shrager and Langley
1990).  In effect, in many disciplines to date discovery is considered quite a
respectable object of investigation.

Contemporary linguistics, unfortunately, did not follow the general
tendency in the other sciences. In the next section, I briefly discuss two
major reasons for this state of affairs, arguing that more attention needs to
be paid to discovery in linguistics.  Then, J.\ S. \mbox{Mill's} methods
for induction are introduced, and a system incorporating these
methods is described, using as  illustrations a simple deciphering problem and
the discovery of (a part of) the famous Germanic Consonant Shift, known as
Grimm's Law.

\section{The Problem of Discovery  in Linguistics}

\subsection{The Transformationalists' Critique of Discovery Procedures}

The study of discovery in linguistics is not fashionable today.  Discovery has
had its good days, reaching its climax in the works of American descriptivists,
and esp.  Zellig Harris (Harris 1951).  The heritage from descriptivists,
however, is by no means uncontroversial.  It is indeed true that their
``discovery procedures'' (of a segmentation-and-classification type),
purporting to uncover the grammar of a language from a corpus of that language,
significantly contributed to the understanding of the process, and served as a
basis for later grammar learning systems and toolkits for linguistic fieldwork.
However, the descriptivists' reduction of linguistic {\em theory} to a manual
of procedures, and doing linguistics to following these procedures proved to be
a too extreme and simplistic view to act as an incentive for the further study
of discovery issues by later generations of linguists.

The attempts of descriptivists, even if somewhat one-sided, suffered a severe
blow with the advent of transformationalists.  In his influential book {\em
Syntactic Structures} Chomsky made a devastating criticism of descriptivists'
discovery procedures, totally shifting the focus on grammar justification.  He
stated of linguistic theory that its ``fundamental concern...is the problem of
{\em justification} of grammars'' (Chomsky 1957: 49; italics mine); and
``we shall never consider how one might have arrived at the grammar'', whether
this be ``by intuition, guess-work, all sorts of partial methodological hints,
reliance on past experience, etc.'' (op.  cit.  p.  56).

Chomsky expressed his doubt as to the attainability {\em in principle} of the
discovery task by descriptivist techniques:

\begin{quotation}

I think that it is very questionable that this goal is attanable in any
interesting way, and I suspect that any attempt to meet it will lead into a
maze of more and more elaborate and complex analytic procedures that will fail
to provide answers for many important questions about the nature of linguistic
structure. (pp. 52-53).\\
...it is questionable whether /procedures of analysis/ {\em can be formulated
rigorously, exhaustively and simply enough to qualify as practical and
mechanical discovery procedure}. (p. 56; italics mine).

\end{quotation}

He also objected to some {\em concrete} attempts in this direction, arguing
that despite their proclaimed goal, they are not in fact discovery, but rather
``evaluation procedures'', helpful for choosing from among alternative
grammars,
already discovered in some way or another (p. 52, fn.3).

Chomsky's disciples followed suit. Another influential transformationalist,
Dougherty, in a review article on linguistic methodology simply repeated
Chomsky's words:

\begin{quote}

I have nothing to say about the creative process by which a linguist develops a
new grammar, I am only concerned with the method of selecting the superior
grammar from a given field of proposed grammars. (Dougherty 1973: 435).

\end{quote}
and Teeter 1964 comments on the question in an article with the
indicative title {\em Descriptive linguistic in America:  triviality vs.
irrelevance}, to mention but a few of the published reactions.

Thus, more than 20 years after Popper, and in phrasing closely reminiscent of
the former, Chomsky achieved an effect in linguistics very much the same as the
one achieved by Popper in philosophy (cf.  fn.1); but linguistics, unlike
philosophy, never fully recovered from the blow.  Not only discovery rules, as
conceived by descriptivists, but the mere word ``discovery'' have eversince
acquired strongly negative connotations for the influential transformational
grammarians, so that many other linguists have had to be very diplomatic on the
subject.  This holds even for some outstanding linguists, ouside the
transformational school.

Thus, some linguists with continuing methodological interests have taken much
care to divert an eventual suspicion that their concerns have anything to do
with discovery, claiming their work to fall entirely into the line of
justification; cf. e.g. Leech 1970, Labov 1971:413-414.

Curiously, attempts have been made to re-interpret the work of the very
proponents of discovery, the descriptivists, in the line of justification,
e.g. Lyons 1970, Miller 1973, Sampson 1979, just in order to rehabilitate them
in the
hypersensitive eyes of Chom\-sky\-ans.  Taking for granted that a concern with
discovery is sinful, it was claimed that:

\begin{quote}

...it is undeniable that the lea\-ding the\-o\-rists /as Bloom\-field,
Ha\-rris, Hockett,
Wells/ (with the exception of Pike) were not concerned with the development of
discovery procedures.\\
...in the work of these linguistis a distinction is carefully drawn between the
actual process of discovering the structure of a language and the business of
describing a structure which has already been discovered.
(Miller 1973: 123).

\end{quote}

It was only the minor representatives of the school, then, who could be charged
with being friends of discovery:

\begin{quote}

Whereas the four linguists cited in the preceding paragraph were the leading
theoreticians of the structuralist school, there were many linguists, less
theoretically minded, who did interpret these techniques of segmentation as
discovery techniques for use in the field. (Miller 1973: 125).

\end{quote}

\subsection{Naive Inductivism}

Another major factor contributing to the neglect of the study of discovery
issues in linguistics is a view which may be called ``naive inductivism''.
What I have in mind is not some worked out system of beliefs, or a coherent
theory, but rather some disparate and vague sentiments, sometimes deeply rooted
in linguistic conscience, as to the primary and exclusive role of ``data
gathering'' and ``observation/generalization'' in the process of linguistic
discovery.  The common implication of all these sentiments is the denial of the
existence of any systematic rules for discovery.

Below I mention two common embodiments of naive inductivism, briefly revealing
their weaknesses.

\subsubsection{The Immediate-Induction-of-Hypothesis Belief}

This amounts to looking upon the process of discovery of a hypothesis (or
problem solution) as springing immediately---and without appeal to any
systematic modes of reasoning---from ``observation'' and ``generalization'' of
the data gathered.

This form of naive inductivism, however, faces serious difficulties (in fact
well-known from the writings of philosophers like Popper and Nagel, among
others).

\begin{itemize}
\item
First, mere observation or data gathering---without some prior
hypothesis/problem---is a poor starting point for making a discovery since we
do not know just which facts to observe or gather.  What we need is the {\em
relevant} facts, which, obviously, presupposes that we already dispose of a
hypothesis/problem.
\item
Secondly, generalizing from facts is (or at least, may be) an activity which is
strictly rule-governed, quite the opposite of what the linguistic proponents of
this view suppose.
\item
And, thirdly, from a body of data not just one, but, as a rule, innumerable
inductive generalizations can be made, so that we need to employ
{\em plausibility} considerations to constrain the choice, and this, again, is
liable to rules.
\end{itemize}

\subsubsection {The Large-Quantity-of-Data Belief}

This sentiment assumes that what facts perhaps cannot do, {\em many} facts can,
and can be traced in linguistic re\-marks, with markedly positive connotation,
to the effect that someone ``is true to the facts'' or ``bases his/her analysis
on large corpuses of data'', etc.  Conversely, other linguists are ridiculed
for for ``having analysed three sentences and a half''.  Consider also the
following quote from an authorative source on the history of linguistics:

\begin{quote}

There are periods in the history of linguistics which look very much like
revolutions and sudden shifts of paradigm, but in my opinion {\em the most
striking aspect of our science is the gradual accumulation over the century of
an immense knowledge about language}...To become aware of this may perhaps be
one of the most significant revolutions in linguistics.
(Hovdhaugen 1982: 11; italics mine).

\end{quote}

This form of inductivism is also very vulnerable.

\begin{itemize}
\item
First, and this point seems quite obvious, the great bulks of data in
themselves are not only not conducive to making a discovery, but are also of a
significant obstacle to it, the selection of the {\em relevant} facts becoming
a more difficult task with the increase of these facts.
\item
And, in the second place, any (linguistic) discovery, contrary to the
tenet
discussed, is as a rule {\em empirically underdetermined}. Norbert Wiener, for
instance, has wittily described the situation. To the question ``On how many
instances would you be willing to base a generalization?'' he is reported to
have answered ``Two instances would be nice, but one is enough!'' (cited in
(Wartowski 1980: 6).
\end{itemize}

Concluding this section, we should note that despite the marked tendency in
current linguistics to disregard, or even be hostile to, the problems of
discovery some linguists remained outside the mainstream (cf.  esp.
Botha 1980 who devotes a whole chapter to a (philosophically-oriented)
treatment of linguistic discovery).  However, there is clearly a need for more
investment of effort.  This paper is a contribution to this trend, but focuses
on computational implementation (cf.  also Pericliev 1990) where some
heuristics are illustrated with a real research problem; a book on linguistic
discovery is under preparation).

\section{Mill's Methods for  Induction}

Summarizing the well-known ideas of the ``Experimental Science'' of the
philosopher F.  Bacon, J.\ S. \mbox{Mill} (Mill 1879) formulated several
methods (``canons'') for discovery of ``causally'' related facts in a set of
instances (observations).  \footnote {On a common understanding causality is
the logical relation ``If A, then B'', where the antecedent A is called a
``cause'' and the consequent B is called an ``effect'', e.g. Harr\'e.
Saying that the cause A has an effect B (or that A causes B) then means simply
that A is an invariable antecedent of B, or equivalently, that B occurs
whenever A occurs.}

Mill's methods assume that each observation con\-sists of a set of putative
causes (or ``accom\-pa\-ny\-ing facts/cir\-cum\-stan\-ces'') for an effect,
their
aim, as eliminative induction methods, being to eliminate, from the set of
putative causes, all but the ``actual'' one(s).

Below we state Mill's heuristics in his own formulation.  Then, the heursitics
are provided with somewhat simplistic linguistic examples, and their
implications for the computational modeling of (linguistic) discovery are
briefly discussed.

\subsection{The Methods}

In the following, ``$\rightarrow$'' means ``accompanies'', ``$\Rightarrow$''
mea
ns
``causes'', ``$\Leftrightarrow$'' means ``either causes or is an effect of'';
ca
pital
letters denote circumstances, and small-case letters the ``phenomena''
investigated.

\begin{itemize}

\item[(1)]
{\em The Method of Agreement (MA)}.  If two or more instances of the
phenomenon under investigation have only one circumstance in common,
the circumstance in which alone all the instances agree, is the cause (or
effect) of the given phenomenon. Schematically:

\begin{tabular}{lcl}
A,B,C &  $\rightarrow$ & a,b,c \\
A,D,E &  $\rightarrow$ & a,d,e \\
\hline
A  & $\Leftrightarrow$ & a
\end{tabular}

\item[(2)]
{\em The Method of Difference (MD)}.  If an instance in which the phenomenon
under investigation occurs, and an instance in which it does not occur, have
every circumstance in common save one, that one occurring only in the former;
the circumstance in which alone the two instances differ, is the effect, or the
cause, or an indispensable part of the cause, of the phenomenon. Schematically:

\begin{tabular}{lcl}
A,B,C &  $\rightarrow$ & a,b,c \\
B,C   &  $\rightarrow$ & b,c \\
\hline
A  & $\Leftrightarrow$ & a
\end{tabular}

\item[(3)]
{\em The Method of Residues (MR)}.  Subduct from any phenomenon such part as is
known by previous inductions to be the effect of certain antecedents, and the
residue of the phenomenon is the effect of the remaining antecedents.
Symbolically:

\begin{tabular}{lcl}
A,B   & $\rightarrow$ & a,b \\
B     & $\Rightarrow$ & b \\
\hline
   A  & $\Rightarrow$ & a
\end{tabular}

\item[(4)]
{\em The Method of Con\-co\-mi\-tant Va\-ri\-a\-tions (MCV)}. Whatever
phenomenon varies in any manner whenever another phenomenon varies in some
particular manner, is either a cause or an effect of that phenomenon, or is
connected with it through some fact of causation.  Symbolically (an apostrophe
denotes a variation):

\begin{tabular}{lcl}
A,B,C   &  $\rightarrow$  &  a,b,c \\
A',B,C   &  $\rightarrow$ &  a', b,c \\
\hline
A  & $\Leftrightarrow$ & a
\end{tabular}

\end{itemize}

\subsection{Examples}

Assume now that we are given the morpheme decomposition of (English) words, as
well as their decomposition into constituent meanings, and we inquire about the
morpheme-meaning correspondences.

{}From the following observations, by MA, we may infer that ``let'' and
``diminutive'' are causally connected.

\begin{tabular}{lcl}
book,let &  $\rightarrow$ & 'book',diminutive \\
leaf,let &  $\rightarrow$ & 'leaf',diminutive \\
book,let,s & $\rightarrow$ & 'book',diminutive \\
\hline
let  & $\Leftrightarrow$ & diminutive
\end{tabular}

Thus, since only the morpheme ``let'' occurs when the resultant meaning
($=$effect) ``diminutive'' occurs, while the other accompanying facts
(``book'',
``leaf'' or ``s'') vary, we conclude that ``let'' causes (or is an
effect of) ``diminutive''.

The following two are self-explanatory examples of the MD and MR, respectively.

\begin{tabular}{lcl}
book,let & $\rightarrow$ & 'book',diminutive \\
book     & $\rightarrow$ &  'book' \\
\hline
let & $\Leftrightarrow$ & diminutive
\end{tabular}
\par

\begin{tabular}{lcl}
book,let,s   & $\rightarrow$ & 'book',diminutive, plural \\
book         & $\Rightarrow$ & 'book' \\
s            & $\Rightarrow$ & plural \\
\hline
   let & $\Rightarrow$ & diminutive
\end{tabular}

As an illustration of the fourth method Mill proposed, the MCV, consider how
one can infer a causal link between accent and grammatical meaning on the basis
of the obseration that the shift of the accent of a word (as e.g.  in
``cond\`uct'' and ``c\`onduct'') leads to a correspondiong shift of the
grammati
cal
meaning, verbal in the first case, and nominal in the second:  \footnote{The
method is also particularly applicable to quantitative terms.}

\begin{tabular}{lcl}
conduct,accent(u)   &  $\rightarrow$  &  gr-meaning(verb) \\
conduct,accent(o)   &  $\rightarrow$ &   gr-meaning(name) \\
\hline
accent(x)  & $\Leftrightarrow$ & gram-meaning(y)
\end{tabular}

\subsection{Some Implications}

The above methods have several features which make their computational modeling
in a linguistic system of considerable interest:

\subsubsection{Generality}

The methods' {\em generality}, as reflected in their domain- and
subject-independence, makes
them applicable to a wide range of linguistic discoveries in diverse linguistic
fields.  Indeed, it is well known that a substantial part of linguistic
``laws''
are in fact empirical regularities, falling under the general schema
'Forms/meanings of type A correspond to/cause forms/mean\-ings of type B' at
the di\-ffe\-rent linguistic levels.

For instance, one of the founders of modern linguistics characterized
synchronic linguistics as finding form-meaning correspondences: ``In human
speech, di\-ffe\-rent sounds have different meanings.  To study the
coordination of certain sounds with certain meanings is to study language.''
(Bloomfield 1933).  A basic task in diachronic (historical) linguistics is the
study of causation of language change in both sounds and meanings.  The study
of language universals, as initiated by J.  Greenberg, most often amounts to
finding so-called ``implicational universals'', etc.

\subsubsection{Psychological Plausibility}

Mill's methods are {\em psychologically plausible} insofar as
they are simple and perfectly natural reasoning modes.
\footnote{This could be explained by noting that the methods actually yield
demonstrative inferences, only making the assumptions that {\em all}
accompanying circumstances are enumerated and that there is no more than one
cause for an effect; the process of elimination itself is indeed trivial. One
is reminded in this context of the words of Bacon in {\em Novum Organum} that
his Method of discovery of sciences leaves little to the acuteness and strength
of wit, and indeed levels wit and intellect.} So they have been widely used in
the
process of linguistic inquiry.  Their use may be unconscious (often disguised
as specific ``(litmus) tests'', ``discovery procedures'', etc., in fact
directly
based on {Mill's} canons), or it may be conscious (e.g.  J.  Greenberg has
recently attributed the discovery of the famous Verner's law to his use of the
MD).  Psychological plausibility is a further feature which a discovery system
may profitably possess.

\subsubsection{Historical Importance}

Finally, the embedding of {\em historically important} methods into
computational systems may serve as test of the particular ideas underlying
these methods, and more generally, as a test of the challenging idea of the
possibility for a purely ``mechanistic discovery''.

\section{System Overview and  Examples}

MILL is a system incorporating the above heuristics.  It does not make the
minute distinctions Mill assumes in the conclusions of the methods;
\footnote{Cf. e.g.  MD, where the conclusion is ``the effect or the cause, or
an indispensable part of the cause''.}
for user-specified cause-sets and effect-sets in observations it merely
identifies a cause for an effect (or vice versa). \mbox{MILL's} {\em data
representation} is a simple Object-Attribute-Value knowledge structure.  Its
{\em discovery process} is, in essence, an attempt to apply one or more of the
above methods.  The system iterates through the knowledge base, trying to apply
a method.  After each successful application, a further simple
heuristic is used, the {\em Elimination Method (EM)}, known to everyone of us,
which checks in the data base whether a putative cause always produces the same
effect, and if not, rejects the conjecture.  A successful conjecture is
attempted to be proven by a further method, and the result is recorded.  The
system repeats this process until all possibilities are exhausted.

As output, MILL produces two interrelated structures, constituting its
discoveries:

\begin{itemize}
\item[(1)]
a set of the causally related facts;
\footnote{An elementary query to the system is of the form {\tt
causation(Cause,Effect)}, where ``Cause'' and ``Effect'' \mbox{stand} for
arbitrary terms, variables or constants; a particular problem formulation may
be stated in terms of any elementary queries, connected by logical operators.}
\item[(2)] an elementary form of ``explanation'' for the system's reasoning
behaviour, comprising the {\em methods} employed to draw a conclusion, as well
as the {\em data} used to this end.  \end{itemize}

In the present implementation, structure (2) is only useful if the system
is being used as a tool by a linguist as an information helping him/her
evaluate the solutions' degree of plausibility.  Thus, some methods (e.g.  MD)
give more plausible results than others (e.g.  MA); a discovery using both
(i.e.  the Joint Method of Agreement and Difference) \footnote{This is actually
a somewhat different interpretation of the Joint Method than Mill gives.} is
more plausible than such using either taken in isolation, etc.

The operation of the system may be better understood by examining particular
examples.

\subsection{A Simple Deciphering Problem}

A text is given, consisting of six phrases in an unknown language A,  together
with their translational equivalents into another unknown language, B.
\footnote{This problem is adapted from (Zaliznjak 1963: 141).  In fact,
language A is Albanian, and language B, Old Jewish.  The Albanian text is given
in the usual orthography.  For the old Jewish text, the Latin transliteration
of consonantal writing is given.} The task is to find the translational
equivalents of the following two words from language B:  \mbox{\em \v sth\/}
and \mbox{\em hzbwb}.

\begin{tabular}{|rll|}
\hline
No. & CAUSES (lg A)  &  EFFECTS (lg B)\\
\hline
1. &  miz\"e, pi      & y\v sth, zbwb            \\
2. &  miza, pinin     & \v stw, zbwbym           \\
3. &  miz\"e, pinte   & \v sth, zbwb             \\
4. &  mizat, pine     & y\v stw, hzbwbym          \\
5. &  miza, pine      & y\v stw, zbwbym          \\
6. &  miza, pi        & y\v sth, hzbwb           \\
\hline
\end{tabular}

The data representation is Object-Attribute-Value.  The first observation e.g.
will be represented as observation(no(1), cause-set([miz\"e, pi]),
effect-set([y\v sth, zbwb])), meaning that the object ``observation'' has three
attributes, number, cause-set and effect-set with their corresponding values.
The system knows that the order of symbols in the cause-set and effect-set is
insignificant.

%Since we need to find the cor\-res\-pon\-den\-ces in lan\-gu\-age A of two
%spe\-ci\-fic words from language B, we shoud for\-mu\-late the prob\-lem as
%ha\-ving two sub\-goals, as follows:  {\tt ?- causality(Cause1,\v sth) $\land$
%causality(Cause2,hzbwb).} (Varaibles begin with a capital letter, and
%constants with a small-case letter.)

Needing to find the cor\-res\-pon\-den\-ces in lan\-gu\-age A of two
spe\-ci\-fic words from language B, the prob\-lem will be for\-mu\-lated  as
follows:  {\tt ?- causality(Cause1,\v sth) $\land$
causality(Cause2,hzbwb).} (Varaibles begin with a capital letter, and constants
with a small-case letter.)
MILL will try to apply one of the methods, looking first for the cause of

{\em \v sth}. Since this word occurs only in obs. No.~3, no other method
except MR is applicable, and it will choose MR.  Applying MR will invoke a
further subgoal, viz.  that of finding---by some of the heuristics---the cause
of {\em zbwb}, the other word in the effect-set of obs.  No.~3.  {\em zbwb}
occurs in obs.  Nos.~1 and 3 and is provable, by MA, to correspond to {\em
miz\"e}.  This being the case, what remains to be the cause of {\em \v sth},
according to MR, is {\em pinte}.  The Elimination Method does not disconfirm
the conjecture {\em pinte} $\Rightarrow$ {\em \v sth}.  In a perfectly
analogous way MILL will solve the second subgoal, finding the cause of {\em
hzbwb} to be {\em miza}.

\subsection{Grimm's Law}

Consider the following data, compiled from Mincoff (1967: 77), which
obeys the well-known Consonant Shift, familiar under the name ``Grimm's Law''
(1822); The Indo-European sounds (exemplified by the Latin words in the left
column) are taken as the causes of the Germanic sounds (exemplified by the Old
English words in the right column).

\begin{tabular}{|rll|}
\hline
No. & CAUSES  &  EFFECTS \\
\hline
1. &  t,u           &     $\delta$,\=u      \\
2. &  t,r,\=e,s     &     $\delta$,r,e,e    \\
3. &  p,a,t,e,r     &     f,a,$\delta$,e,r   \\
4. &  p,\=e,s       &     f,\=o,t   \\
5. &  p,e,c,u       &     f,e,o,h       \\
6. &  n,e,p,\=o,s   &     n,e,f,a      \\
7. &  d,u,o         &     t,w,\=a      \\
8. &  d,e,c,e,m     &     t,\=e,n      \\
\hline
\end{tabular}

Let the  problem be to discover the {\em consonantal} alternations exhibited in
the data.

To solve the problem MILL will need an explicit encoding of the {\em domain
knowledge} as to what of the symbols in the data base are c(onsonants) and what
are v(owels). The representation, containing this information, taking as an
illustration e.g.  obs.  No.~4 will be
observation(no(4),cause-set([c:p,v:\=e,c:s]), effect-set([c:f,v:\=o,c:t])).

The problem formulation, unputted to the system, will be:

{\tt ?- causation(c:Cause,c:Effect).}

Now we may look briefly at the discovery process.  Trying to apply MA, the
system notices in observations Nos.~1 and 2 the single co-occurring symbol in
their cause-sets,{\em t\/}, and the single co-occurring symbol in their
effect-sets, $\delta$.  \footnote{The same conjecture could also have been made
by the same method from observations Nos.~1 and 3, had the system used
another mode of scanning tha data base than the top-down one.} Then it proceeds
with EM.  This test succeeding, since whenever {\em t} occurs $\delta$ also
occurs (as in obs.  No.~3) and there is no contradicting data in the base, it
is hypothesized that {\em t} causes $\delta$.  This hypothesis is then
attempted to be proven by a further method (failing in this particular case),
and the result is recorded.

By the same method, from observations Nos.~3 and 4, the system will
hypothesize that {\em p\/} causes {\em f\/}, the EM will succeed since in
observations Nos.~ 5 and 6 where {\em p} occurs the conjectured correspondence
{\em f} also occurs, and {\em p} causes {\em f} will be assumed.  Again using
the MA, and simialr reasoning route the system will infer from observations
Nos.~7 and 8 that {\em d} causes {\em t}.

MILL has thus re-discovered the Indo-European--Germanic  consonantal
alternations {\em t~$>$~$\delta$, p~$>$~f, d~$>$~t}, which form a part of the
famous Grimm's Law (1882).  \footnote {Grimm's Law, together with some other
sound laws, like that of Verner, was one of the greatest achievements of 19th
c.  linguistics.}

\section{Conclusion}

At this stage of development, MILL has certain limitations, arising both from
Mill's conception (our interpretation) of causality and the implementation.
To mention but one thing, it is incapable of handling ``exceptions''. E.g.
had we encountered in our data base of the consonantal alternation problem an
observation such as Latin {\em st\=o} O.E. {\em standan}, MILL would have
ne\-ver found the change {\em t~$>$~$\delta$}. Nevertheless, it will be
clear
that a discovery process, relying on Mill's methods, will make a system simple,
and at the same time quite general and powerful; thus the system has
re-discovered a number of further sound laws, and was successfully tested in
its role as a tool in the solutions of diverse ``field'' linguistic problems.
Finally, although we have considered only linguistic discovery, the system may
be, obviously, applied to discovering empirical regularities in other fields as
well.

\end{document}